\documentclass{article}

     \usepackage[preprint,nonatbib]{neurips_2020}

     \usepackage[numbers]{natbib}

\usepackage[utf8]{inputenc} 
\usepackage[T1]{fontenc}    
\usepackage{hyperref}       
\usepackage{url}            
\usepackage{booktabs}       
\usepackage{amsfonts}       
\usepackage{nicefrac}       
\usepackage{microtype}      

\usepackage{amssymb,amsmath,amsthm,dsfont,bm,mathtools}
\usepackage{color,graphicx,epsfig}
\usepackage[capitalise]{cleveref}
\usepackage{multirow,bigstrut,rotating}
\usepackage{capt-of,caption,subfig}
\usepackage{algorithm}
\usepackage[noend]{algpseudocode}
\usepackage{wrapfig}
\usepackage[title]{appendix}
\usepackage[multiple]{footmisc}

\usepackage{enumitem}
\setlist{topsep=0pt, partopsep=0pt, leftmargin=13pt}

\usepackage{xcolor}
\hypersetup{
	colorlinks,
	linkcolor={red!50!black},
	citecolor={blue!50!black},
	urlcolor={blue!80!black}
}

\newcommand{\loss}{{L}}
\newcommand{\yhat}{{\hat{y}}}
\newcommand{\xtilde}{{\tilde{x}}}
\newcommand{\ytilde}{{\tilde{y}}}
\newcommand{\ybar}{{\bar{y}}}

\newcommand{\what}{{\hat{w}}}

\newcommand{\smplst}{{\mathcal{S}}}

\newcommand{\reg}{{R}}
\newcommand{\dist}{{p}} 
\newcommand{\conf}{{\tau}}

\newcommand{\fbase}{{f_{\text{base}}}}
\newcommand{\fla}{{f_{\text{look}}}}
\newcommand{\mask}{{\Gamma}}
\newcommand{\dec}{{d}}
\newcommand*\diff{\mathop{}\!\mathrm{d}}
\newcommand{\losspred}{{\loss_{\mathrm{pred}}}}
\newcommand{\lossuncert}[1]{{\loss^{({#1})}_{\mathrm{uncert}}}}
\newcommand{\lossprop}{{\loss_{\mathrm{prop}}}}

\newcommand{\red}[1]{{\leavevmode\color{red}{#1}}}
\newcommand{\blue}[1]{{\leavevmode\color{blue}{#1}}}

\newcommand\todo[1]{\red{TODO: {#1}}}
\newcommand\tocite{\red{[CITE]}}

\newcommand{\beq}{\begin{equation}}
\newcommand{\eeq}{\end{equation}}
\newcommand{\bal}{\begin{align}}
\newcommand{\eal}{\end{align}}
\newcommand\expect[2]{\mathbb{E}_{#1}{[ {#2} ]}}
\newcommand\prob[2]{\mathbb{P}_{#1}{\left[ {#2} \right]}}

\DeclareMathOperator*{\argmin}{argmin}

\newcommand{\1}[1]{\mathds{1}{\{{#1}\}}}

\newcommand{\naive}{{na\"{\i}ve}}

\newcommand{\X}{{\cal{X}}}

\newcommand{\R}{{\mathbb{R}}}

\newtheorem{assumption}{Assumption}

\title{From Predictions to Decisions: \\ Using Lookahead Regularization}

\author{%
Nir Rosenfeld \\
  School of Enginreering and Applied Science\\
  Harvard University\\
  \texttt{nirr@g.harvard.edu}\\
   \And
   Sophie Hilgard\\
  School of Enginreering and Applied Science\\
Harvard University\\
\texttt{ash798@g.harvard.edu}\\
   \AND
   Sai Srivatsa Ravindranath\\
  School of Enginreering and Applied Science\\
Harvard University\\
\texttt{saisr@g.harvard.edu}\\
   \And
   David C. Parkes\\
  School of Enginreering and Applied Science\\
Harvard University\\
\texttt{parkes@eecs.harvard.edu}\\
}

\begin{document}

\maketitle

\begin{abstract}

Machine learning is a powerful tool for predicting  human-related outcomes,
from credit scores to heart attack risks.
But when deployed, learned models also affect how users act 
in order to improve outcomes, whether predicted or real.
The standard approach to learning
is agnostic to induced user actions
and provides no guarantees as to the effect of  actions.
We provide a framework for learning predictors that  are both  accurate and promote good actions.
For this, we introduce  {\em look-ahead regularization} which,
by anticipating user actions, encourages predictive models to also induce actions that improve outcomes.
This regularization carefully tailors the uncertainty estimates governing confidence in this improvement  to the distribution of model-induced actions.
We report the results of experiments on real and synthetic data that
show the effectiveness of this approach.

\end{abstract}


\section{Introduction} \label{sec:intro}

Machine learning is increasingly being used in domains that have considerable impact on people, ranging from healthcare \cite{callahan2017machine} to banking \cite{siddiqi2012credit} to manufacturing \cite{wuest2016machine}. Moreover, in many of these domains, the desire for transparency has led to published machine-learned models that play a dual role in prediction and influencing behavior change. Consider a doctor who uses a risk tool to  predict whether a patient is in danger of having a heart attack, while at the same time wanting to recommend lifestyle changes to improve outcomes.\footnote{MDCalc is one example of a site that provides risk assessment calculators for use by medical professionals \url{https://www.mdcalc.com/}} Consider a bank, looking to predict whether customers are likely to repay loans while  customers are, at the same time, seeking to improve their credit scores. Consider a wine producer, looking to predict the demand for a new vintage while at the same time deciding how to make changes to their production process to improve future vintages.

It is well understood that correlation and causation need not go hand-in-hand \cite{pearl2009causal, rubin2005causal}. 
What is novel about this work is that we 
seek models that serve the dual purpose of achieving predictive accuracy 
as well as 
providing high confidence
that decisions made with respect
to the model improve outcomes.
That is, we care  about the  utility that comes from  having a predictive tool, while recognizing that these tools may 
also drive decisions.
To illustrate the potential pitfalls
of a purely predictive approach, 
consider a doctor who would like to advise a patient on how to reduce  risk of heart attack.
If the doctor assesses risk using a linear predictive model (as is often the case, see \cite{ustun2016supersparse}),
then a negative coefficient for alcohol consumption may lead the doctor to suggest a daily glass of red wine. Is this decision justified?
Perhaps not, although this recommendation has often been made based on correlative evidence and despite a clear lack of experimental support \cite{haseeb2017wine, sahebkar2015lack}.

At the same time,
predictive models are valuable in and of themselves, 
for example in assessing whether a patient is in immediate risk.
%
%
%
%
Similarly, banks want to understand credit risk while promoting good decisions by consumers in regard to true 
creditworthiness, and wine producers want to predict the marketability of a new vintage while improving their processes for next
year.
As designers of a learning framework, what degrees of freedom can we utilize
to promote good decisions?
Our main insight is that 
{\em controlling the tradeoff between accuracy and decision quality,
where it exists, can be cast as a problem of model selection}.
For instance,  there may be multiple models  with similar 
predictive performance
but with different coefficients that therefore induce very different decisions \cite{breiman2001statistical}.
%
%
To achieve this
tradeoff we introduce {\em  lookahead regularization}, which   balances   accuracy and the improvement associated with induced decisions. 
 Lookahead regularization anticipates how  users will act  and penalizes a model unless there is high confidence that 
these decisions will  improve  outcomes. 

Formally, these decisions, which depend on the predictive model, induce a target distribution $\dist'$ on covariates
that may differ from an initial  distribution $\dist$. 
For an individual with covariates $x$, they are mapped to new covariates $x'$.
%
For a prespecified confidence level $\conf$,  we   want to 
guarantee improvement for at least a $\tau$-fraction of the population, 
comparing outcomes under $\dist'$ in relation to observed outcomes in $\dist$ (under an invariance assumption on $p(y|x))$.
%
%
%
%
%
The  technical challenge 
is that 
 $\dist'$ may differ considerably from $\dist$,
resulting in uncertainty in  estimating  the effect of decisions.
To solve this, lookahead regularization makes use of an uncertainty
model that provides  confidence intervals around decision outcomes.
A discriminative uncertainty model
 is trained under a learning framework that 
makes use of importance weighting~\cite{gretton2009covariate,shimodaira2000improving,sugiyama2008direct}
%
%
to handle covariate shift 
and is designed to provide accurate intervals  for $\dist'$.
%


%
Our  algorithm  has stages that alternate between optimizing the different components
of our framework: the {\em predictive model} (under the lookahead regularization term),
the {\em uncertainty model} (used within the regularization term),
and the {\em propensity model} (used for covariate shift adjustment).
If the uncertainty model is differentiable and the predictive model is twice-differentiable,
then gradients can pass through the entire pipeline and gradient-based optimization can be applied.
We run three experiments. One uses synthetic data and  illustrates how our approach can be useful,
as well as helping to understand what is needed for it to succeed. The second application
is to  wine quality prediction and shows that even  simple  tasks have interesting tradeoffs
between accuracy and improved decisions that can be utilized.
The third experiment focuses on predicting diabetes progression
and includes a demonstration of the framework 
in a setting with individualized actions.


\subsection{Related work} \label{sec:related}

\textbf{Strategic Classification.}
In the field of \emph{strategic classification}, the learner and agents engage in a Stackelberg game, where the learner attempts to publish a maximally accurate classifier 
taking into account that agents will shift their features to obtain better outcomes under the classifier
\cite{hardt2016strategic}.
While early efforts viewed all modifications as ``gaming"--- an adversarial effect to be mitigated \cite{dong2018strategic,bruckner2011stackelberg} ---a recent trend has focused on 
creating incentives 
for modifications that lead to better outcomes \emph{under the ground truth function} rather than simply better classifications \cite{kleinberg2019classifiers, alonmultiagent,  haghtalab2020maximizing,tabibian2019optimal}. In the absence of a known mapping from effort to ground truth, \citet{miller2019strategic} show that incentive design relates to causal modeling, and several responsive works explore how the actions induced by classifiers can facilitate discovery of these causal relationships \cite{perdomo2020performative,bechavod2020causal,shavit2020learning}. The second order effect of strategic classification on algorithmic fairness has  also motivated several works \cite{liu2020disparate,hu2019disparate,milli2019social}.
Generally, these works consider the equilibrium effects of classifiers, where the choice of model affects covariate distributions and in turn predictive accuracy. In contrast, we consider what can be done given a snapshot at a point in time, or when the input distribution remains unaffected by user actions.

\textbf{Causality, Covariate Shift, and Distributionally Robust Learning.} There are many efforts in ML to quantify the uncertainty associated with predictions and identify domain regions where models err \cite{lakshminarayanan2017simple, hernandez2015probabilistic, gal2016dropout, guo2017calibration, tagasovska2019single, liu2019accurate}.
However, most methods fail to achieve desirable properties when deployed out of distribution (OOD) \cite{snoek2019can}.
When the shifted distribution is unknown at train time,
distributionally robust learning can provide worst-case guarantees for
specific types of shifts but require unrealistic computational expense or restrictive assumptions on model classes \cite{sinha2017certifying}.
Although we do not know ahead of training our shifted distribution of interest,
our framework is concerned only with the single, specific 
OOD distribution that is induced by the learned predictive model.
Hence, we need only guarantee robustness to this particular distribution,
for which we make use of tools from learning under covariate shift \cite{bickel2009discriminative}.
Relevant to our task, \citet{mueller2016learning} seek to identify treatments which are beneficial with high probability under the covariate shift assumption.
Because model variance generally increases when covariate shift acts on non-causal variables \cite{peters2016causal}, our framework of trading off uncertainty minimization with predictive power relates to efforts in the causal literature to find models which have optimal predictive accuracy while being robust to classes of interventional perturbations \cite{meinshausen2018causality, rothenhausler2018anchor}.

\begin{figure}[t]
	\centering
	\includegraphics[width=1\textwidth]{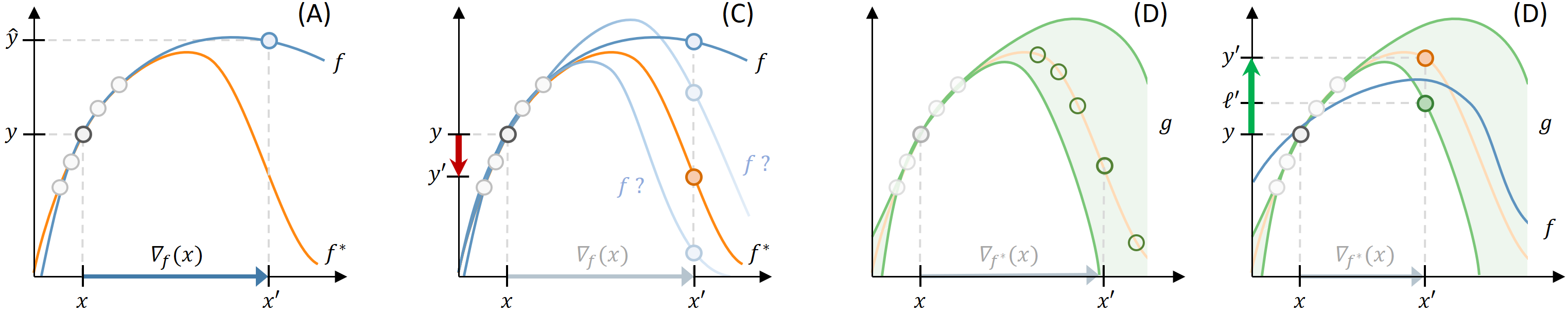}
	\caption{An illustration of our approach.
		The data density $\dist(x)$ is concentrated to the left of the peak,
		and $\dist(y|x)$ is deterministic, $y=f^*(x)$.
		(A) Users ($x$) seeking to improve their outcomes ($y$)
		often look to predictive models for guidance on how to act,
		e.g. by following gradient information ($x \mapsto x'$).
		(B) But actions may move $x'$ into regions of high uncertainty
		where $f$ is unconstrained by the training data,
		and models of equally good fit on $\dist$ can behave very differently on $\dist'$.
		(C) To reason about the uncertainty in decision outcomes,
		our approach learns an interval model $g(x')=[\ell',u']$,
		decoupled from $f$ and targeted specifically at $\dist'$,
		guaranteeing that $y' \in [\ell',u']$ with confidence $\conf$.
		(D) By incorporating into the objective a model of user behavior,
		lookahead regularization (Eq. \eqref{eq:la_reg}) 
		allows for balancing between accuracy and improvement.
		By penalizing $f$ whenever $y>\ell'$,
		the model learns predictive models that encourage decisions that are safe,
		i.e., $y'\ge y$ w.p. at least $\tau$.
	}
	\label{fig:illust}
\end{figure}

\section{Method} \label{sec:method}

Let $x \in \X=\R^d$ denote a feature vector and $y \in \R$ denote a label,
where $x$ describes the object of interest (e.g., a patient, a customer, a wine vintage), and 
$y$ describes  the quality of an outcome associated with $x$, where 
we assume that higher $y$ is better.
%
%
We assume an observational dataset  $\smplst=\{(x_i,y_i)\}_{i=1}^m$,
which consists of IID samples from a 
population with joint distribution $(x,y)\sim\dist(x,y)$
over covariates (features) $x$ and outcomes $y$.
We denote by $\dist(x)$ the marginal  distribution
on covariates.

Let $f:\X\rightarrow \R$ denote a model
trained on $\smplst$.  We assume that $f$ is used
in two different ways:
\begin{enumerate}
\item \textbf{Prediction:} To predict outcomes $y$ for objects $x$, sampled from $\dist(x)$.
\item \textbf{Decision:} To take action, through changes to $x$, with the goal of improving outcomes.
\end{enumerate}
We will assume that user actions map each $x$ to a new $x' \in \X$.
We refer to $x'$ as a user's \emph{decision} or \emph{action}
and denote decision outcomes by $y' \in \R$,
We set $x'=\dec(x)$ and refer to $\dec : \X \rightarrow \X$ as the \emph{decision function}.
We will assume that users consult $f$ to drive decisions---either because they care only about 
predicted outcomes (e.g., the case of bank loans),
or because they consider the model to be a valid proxy  of the effect
of a decision on the outcome 
(e.g., the case of heart attack risk or wine production).
As in other works incorporating strategic users into learning \cite{perdomo2020performative,hardt2016strategic},
our framework requires an explicit model of how users use the model to make decisions.
For concreteness, we model users as making a step
in the direction of the gradient of $f$,
but note that the framework can also support any other differential decision model.\footnote{Many works consider decisions that take gradient steps under cost constraints $c(x,x') < B$.
	Note that such constraints can be incorporated into learning using, for example,
	differentiable optimization layers \cite{agrawal2019differentiable}.}
Since not all attributes may be susceptible to change,
we distinguish between {\em mutable} and {\em immutable} features using a \emph{masking operator}
$\mask:\X\rightarrow \{0,1\}^d$.
\begin{assumption}[User decision model]
	\label{ass:user_model}
	Given masking operator $\mask$, we define
	user decisions as:
	\begin{equation}
	x' = x + \eta \mask(\nabla_f(x)),
	\end{equation}
	where the {\em step size} $\eta \ge 0$ is a design parameter.
\end{assumption}

Through Assumption~\ref{ass:user_model}, user decisions induce
a particular decision function $\dec(x)$, and in turn,
a \emph{target distribution} over $\X$, which we denote $\dist'(x)$.
This leads
to a new joint distribution $(x',y')\sim \dist'(x,y)$, 
with decisions inducing new outcomes.
To achieve causal validity in the way we reason about the effect of decisions on outcomes,
we follow~\citet{peters2016causal}
and assume
that $y$ depends only on $x$ and is invariant to the distribution on $x$.
\begin{assumption}[Covariate shift \cite{shimodaira2000improving}] \label{ass:cov_shift}
	The conditional distribution on outcomes, $p(y|x')$, is 
	invariant for any arbitrary, marginal distribution $\dist'(x)$
	on covariates, including the data distribution $\dist(x)$.
\end{assumption}

Assumption~\ref{ass:cov_shift} says that whatever the transform $\dec$,
conditional distribution $\dist(y|x)$ is fixed, and
the new joint distribution is 
$\dist'(x',y)=\dist(y|x')\dist'(x')$, for any $\dist'$
(note that $\dist'$ also depends on $f$).
This covariate-shift assumption ensures the causal validity of our approach
(and entails the property of {\em no-unobserved confounders}).
Although  a strong assumption, this kind of invariance has been leveraged
in other works that relate to questions of causality~\cite{rojas2018invariant,mueller2016learning},
as well as for domain adaptation~\cite{schweikert2009empirical,quionero2009dataset}. 


%

\subsection{Learning objective}

%
Our goals in designing a learning framework 
are twofold.
First, we would like learning to result in a model whose predictions $\yhat=f(x)$
closely match the corresponding labels $y$ for $x \sim \dist(x)$.
Second,
we would like the model to induce  decisions $x'$ for counterfactual distribution $\dist'$
whose outcome $y'$ improves upon the initial $y$.
%
%
To balance between these two goals,  we  construct a learning objective in which a predictive
loss function  is augmented with a regularization term
that promotes good decisions. 
The difficulty is that decision outcomes $y'$
depend on decisions $x'$ through the learned model $f$.
Hence, realizations of $y'$ are unavailable at train time,
as they cannot be observed  until after the model is deployed.
For this reason, simple constraints of the form $y' \ge y$ are  ill-defined, 
and to regularize we must reason about  outcome distributions $y' \sim \dist(y|x')$,
for $x'\sim \dist'$. 
%
A naive approach might  consider the average improvement,
with
$\mu' = \expect{y'\sim\dist(y|x')}{y'}$, for a given $x'\sim \dist'$,
and penalize the model whenever $\mu' < y$, for
example linearly using $\mu'-y$.
Concretely, $\mu'$ must be estimated, and since $f$ minimizes MSE, then 
 $\yhat'=f(x')$ is a plausible estimate of $\mu'$, giving:
\beq
\label{eq:objective_naive}
\min_{f \in F} \expect{\dist(x,y)}{(\yhat-y)^2} + \lambda \expect{\dist(x,y)}{\yhat'-y},
\qquad \yhat'=f(x'),
\eeq
where $\lambda\geq 0$ determines the relative importance of improvement over accuracy.
%
There are two issues with this approach.
First, learning can result in an $f$ that severely overfits in estimating $\mu$,
meaning that at train time the penalty term in the (empirical) objective will appear to be low
whereas at test time its (expected) value will be high.
This can happen, for example, when $x'$ is moved to a low-density region of $\dist(x)$
where $f$ is unconstrained by the data and, if flexible enough,
can artificially (and wrongly) signal improvement. 
To address this we use two decoupled models---one for predicting $y$ on
distribution $\dist$,
and another for handling $y'$ on distribution $\dist'$.

Second,
in many applications it may be unsafe to guarantee that improvement hold only on average
per individual (e.g., heart attack risk, credit scores). 
To address this, we encourage $f$ to improve
 outcomes with a certain degree of confidence $\conf$, for $\conf>0$,
i.e., such that $\prob{}{y' \ge y} \ge \conf$ for a given $(x,y)$ and induced $x'$ and thus $p(y'|x')$.
Importantly, while one source of uncertainty in $y'$ is $\dist(y|x')$,
other sources of uncertainty exist, including those coming from insufficient data 
as well as  model uncertainty.
Our formulation is useful when additional sources of uncertainty are significant,
such as when the model $f$ leads to actions that place $x'$ in low-density regions of $\dist$.

In our method, we 
replace the average-case penalty in Eq. \eqref{eq:objective_naive}
with a  {\em confidence-based penalty}:
\begin{equation}
\label{eq:conf_objective}
\min_{f \in F} \expect{\dist(x,y)}{(\yhat-y)^2} + \lambda
\expect{\dist(x,y)}{\1{\prob{}{y' \ge y} < \conf}},
\qquad y' \sim \dist(y|x'),
\end{equation}
where $\1{A}=1$ if $A$ is true, and $0$ otherwise. 
In practice, $\prob{}{y' \ge y}$ is unknown, and must be estimated.
For this, we make use of an {\em  uncertainty model}, $g_\conf:\X\rightarrow \R^2$, $g_\conf \in G$,
which we  learn, and  maps points $x'\in\X$ to intervals $[\ell',u']$ that cover 
$y'$ with probability $\conf$.
We also replace the penalty term in Eq. \eqref{eq:conf_objective}
with the slightly more conservative $\1{\ell' < y}$,
and to make learning feasible 
we  use the hinge loss $ \max\{0, y-\ell'\}$ as a convex surrogate.\footnote{The penalty is conservative in that it considers only one-sided uncertainty, i.e., $y' < \ell$ and $u$ is not used explicitly.
	Although open intervals suffice here,
	most methods for interval prediction consider closed intervals,
	and in this way our objective can support them.
	For symmetric intervals, $\conf$ simply becomes $\conf/2$.}
For a given uncertainty model, $g_\conf$,
the empirical learning objective for model $f$ on  sample set 
$\smplst$ is:
\begin{equation}
\min_{f \in F} \sum_{i=1}^m {(\yhat_i-y_i)^2} + \lambda \reg(g_\conf;\smplst),
\quad \mbox{where} \ \ 
\reg(g_\conf;\smplst) = \sum_{i=1}^m {\max\{0, y_i-\ell'_i\}},
\label{eq:la_reg}
\end{equation}

where $\reg(g_\conf;\smplst)$ is the \emph{lookahead regularization} term. 

By anticipating how users decide, 
this penalizes models whose induced decisions do not improve outcomes at a sufficient rate
(see Figure \ref{fig:illust}).
%
The novelty in the regularization term  is that it accounts for uncertainty in assessing improvement,
and does so for points $x'$ that are out of distribution. If $f$ pushes $x'$ towards regions of high uncertainty,
then the interval $[\ell',u']$ is likely to be large,
and $f$ must make more ``effort" to guarantee improvement, something that may  come at some cost to in-distribution prediction accuracy.
As a byproduct, while the objective encodes the rate of decision improvement,
we will also see the magnitude  of improvement 
increase in our experiments.

Note that the regularization term $\reg$ depends both on $f$ and $g$---to determine $x'$, and to determine $\ell'$ given $x'$, respectively.
This justifies the need for the decoupling of $f$ and $g$:
without this, uncertainty estimates based on $f(x')$
are prone to overfit by artificially manipulating intervals to be higher than $y$,
resulting in low penalization at train time without actual improvement (see Figure~\ref{fig:synth} (right)).

\subsection{Estimating uncertainty}

The usefulness of lookahead regularization  relies on the ability of the uncertainty model $g$ to correctly capture the various kinds of uncertainties about the outcome value for the perturbed points.
This can be difficult because uncertainty estimates are needed for out-of-distribution points $x'$.
%

Fortunately, for a given $f$ the counterfactual distribution $\dist'$ is known (by Assumption~\ref{ass:user_model}),
and we can use the covariate transform
associated with the decision to construct sample set $\smplst'=\{x'_i\}_{i=1}^m$.
Even without labels for $\smplst'$, estimating $g$ is now  a problem of 
{\em learning under covariate shift},
where the test distribution $\dist'$ can differ from the training distribution $\dist$.
In particular, 
we are interested in learning uncertainty 
intervals that provide good coverage.
There are many approaches to learning under covariate shift.
Here we describe the simple and popular method 
importance weighting, or inverse propensity weighting
\cite{shimodaira2000improving}.
For a loss function $\loss(g)=\loss(y,g(x))$, we would like to minimize
$\expect{\dist'(x,y)}{\loss}$.
Let $w(x) = \dist'(x) / \dist(x)$, then by the covariate shift assumption:
\begin{equation*}
\expect{\dist'(x,y)}{\loss(g)} = 
\int \loss(g) \diff \dist'(x) \diff \dist(y|x) =
\int \frac{\dist'(x)}{\dist(x)} \loss(g) \diff \dist(x) \diff \dist(y|x) =
\expect{\dist(x,y)}{w(x) \loss(g)}.
\end{equation*}

Hence, training $g$ with points sampled from distribution $\dist$ but weighted by $w$
will result in an uncertainty model that is optimized for
the counterfactual distribution $\dist'$.
In practice, $w$ is itself unknown, but many methods exist for learning an approximate model $\what(x)\approx w(x)$
using sample sets $\smplst$ and $\smplst'$ (e.g. \cite{kanamori2009least}).
To remain within our discriminative approach,
here we follow \cite{bickel2009discriminative} and train a logistic regression model $h:\X\rightarrow [0,1]$, $h \in H$,
to differentiate between points $\xtilde \in \smplst$ (labeled $\ytilde=0$)
and $\xtilde \in \smplst'$ (labeled $\ytilde=1$) and set weights to $\what(x)=e^{h(x)}$.
As we are interested in training $g$ to gain
a coverage guarantee, we define $\loss(y,g(x))=\1{y \notin [\ell,u]}$.

\subsection{Algorithm}

All the elements in our framework--- the predictive model $f$, the uncertainty model $g$, and the propensity model $h$ ---are interdependent.
Specifically, optimizing $f$ in Eq. \eqref{eq:la_reg} requires intervals from $g$;
learning $g$ requires weights from $h$; and $h$ is trained on $\smplst'$ which is in turn determined by $f$.
Our algorithm therefore alternates between optimizing each of these components while keeping the others fixed.
At round $t$, $f^{(t)}$ is optimized with intervals $[\ell'_i,u'_i] = g^{(t-1)}(x'_i)$,
$g^{(t)}$ is trained using weights $w_i = h^{(t)}(x_i)$,
and $h^{(t)}$ is trained using points $x'_i$ as determined by $f^{(t)}$.
The procedure is initialized by training $f^{(0)}$ without the lookahead term $\reg$.
%
%
For training   $g$ and $h$, weights $w_i = \what(x_i)$ and points $\{x'_i\}_{i=1}^m$, respectively,
can be precomputed and plugged into the objective.
Training $f$ with Eq. \eqref{eq:la_reg}, however, requires access to the \emph{function} $g$,
since during optimization, the lower bounds $\ell'$ must be evaluated for points $x'$ that vary
as updates to $f$ are made. Hence, to optimize $f$ with gradient methods,
we use an uncertainty model $g$ that is differentiable, so that gradients can pass through them
(while keeping their parameters fixed). Furthermore, since gradients must also pass through $x'$
(which includes $\nabla_f$), we require that $f$ be twice-differentiable.

In the experiments we consider two methods for learning $g$:
\begin{enumerate}
\item 
Bootstrapping \cite{efron1994introduction},
where a collection of models $\{g^{(i)}\}_{i=1}^k$ is trained for prediction 
each on a subsampled dataset and combined to produce a single interval model $g$, and

\item
Quantile regression \cite{koenker2001quantile},
where models $g^{(\ell)}, g^{(u)}$ are discriminatively trained to
estimate the
$\tau$ and $1- \tau$ quantiles, respectively, of the 
counterfactual target distribution $\dist'(y|x')$.


\end{enumerate}




\begin{figure}[t]
	\centering
	\includegraphics[width=0.32\textwidth, clip]{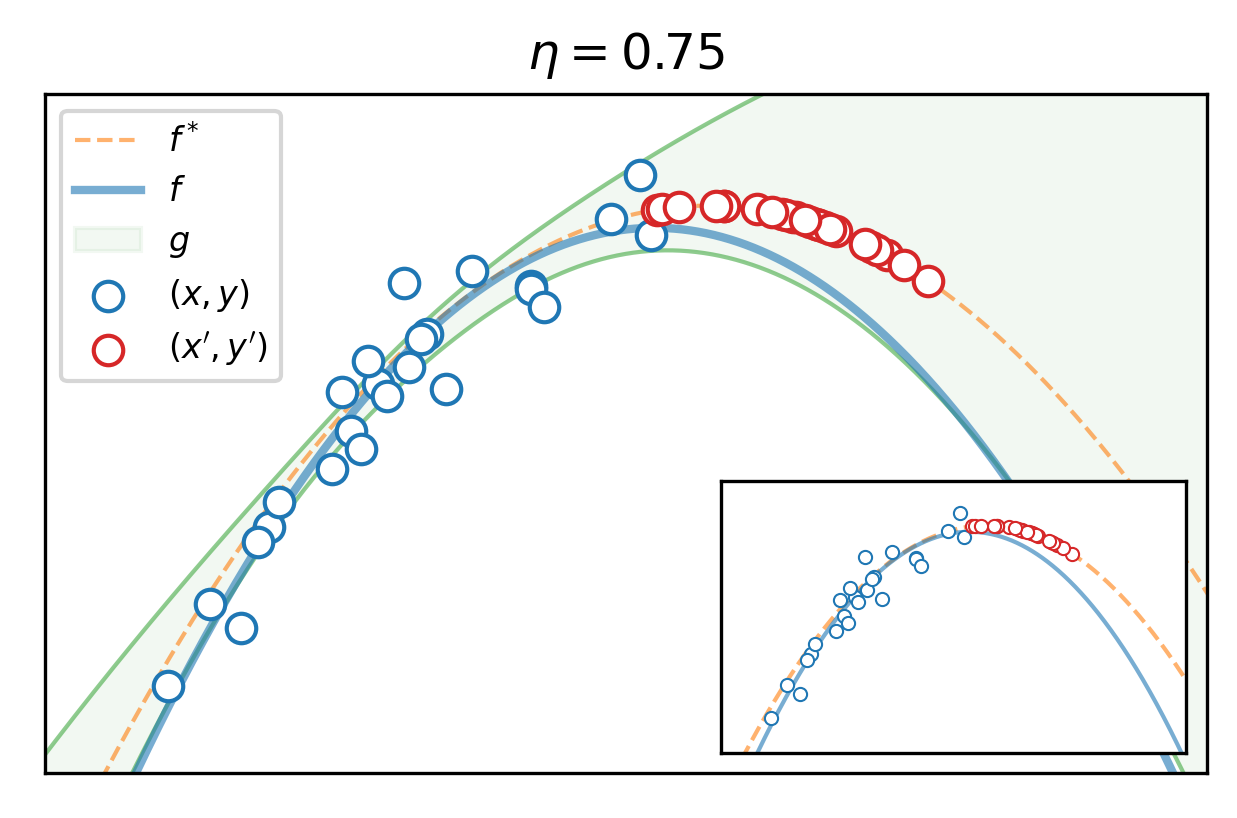}
	\includegraphics[width=0.32\textwidth, clip]{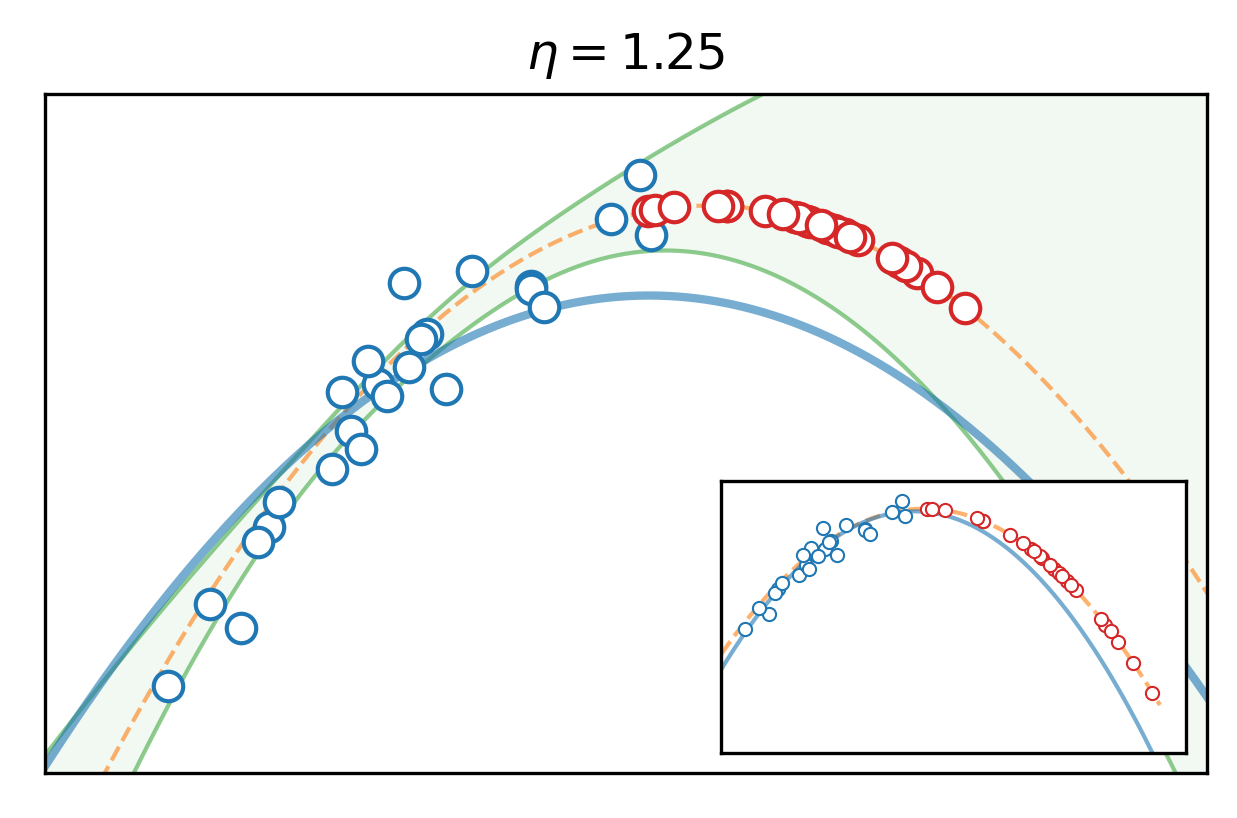}
	\includegraphics[width=0.32\textwidth, clip]{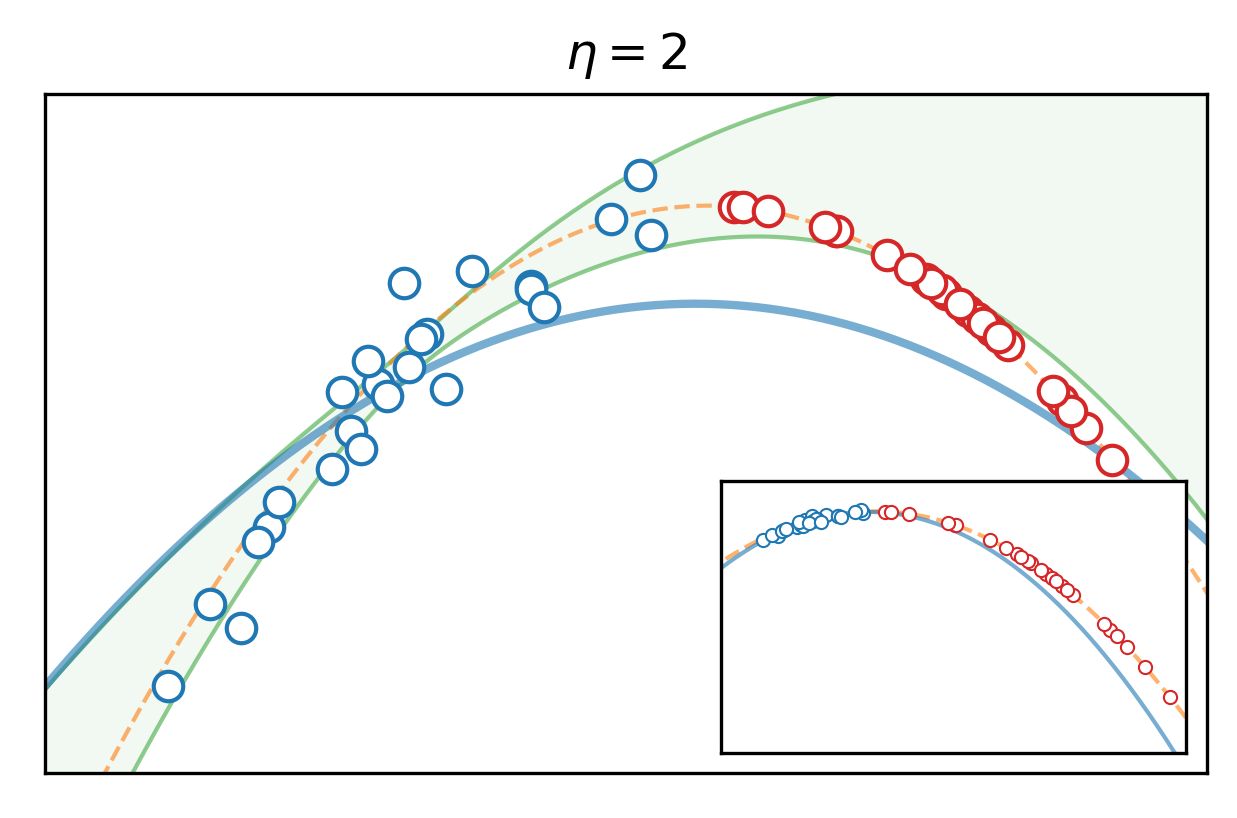}
	\caption{Results for the synthetic experiment for lookahead
	(main plot) and baseline (inlay) models.
	\label{fig:synth}}
\end{figure}

\section{Experiments} \label{sec:experiments}

In this section, we evaluate our approach  in three experiments of 
increasing complexity and scale, where the first is synthetic
and the latter two use real data.
Because the goal of regularization is to balance accuracy with decision quality,
we will be interested in understanding the attainable frontier of accuracy vs. improvement.
For our method, this will mostly be controlled by varying lookahead regularization
parameter, $\lambda\geq 0$.
In all experiments we measure predictive accuracy with root mean squared error (RMSE),
and decision quality in two ways: {\em mean improvement rate}
$\expect{}{\1{y'_i>y_i}}$
and {\em mean improvement magnitude} $\expect{}{y'_i-y_i}$.

To evaluate the approach, we need a means for evaluating counterfactual outcomes $y'$
for  decisions $x'$.
Therefore, and similarly to~\citet{shavit2019extracting}, we
make use of an inferred `ground-truth' function $f^*$ to test decision improvement, 
assuming $y'=f^*(x')$. Model $f^*$ is trained on the entirety of the data.
By optimizing $f^*$ for RMSE, we think of this as estimating the conditional mean of $\dist(y|x)$,
with the data labels as (arbitrarily) noisy observations.
To make for an interesting experiment,
we learn $f^*$ from a function class $F^*$ that is more expressive than $F$ or $G$.
The  sample set $\smplst$ 
will contain a small and possibly biased subsample of the data,
which we call the `active set', and that  plays the role of a representative sample from $\dist$.
This setup allows us not only to evaluate improvement, but also to experiment
with the effects of different
sample sets.


\subsection{Experiment 1: Quadratic curves}

For a simple setting, we explore the effects of regularized and unregularized learning on decision quality
in a stylized setting using unidimensional quadratic curves.
Let $f^*(x)=-x^2$, and assume $y=f(x)+\varepsilon$ where $\varepsilon$ is 
independently, normally distributed.
By varying the decision model step-size $\eta$, we explore three conditions:
one where a \naive\ approach works well,
one where it fails but regularization helps,
and one where regularization also fails.

In Figure \ref{fig:synth} (left), $\eta$ is small,
and the $x'$ points stay within the high certainty region of $\dist$.
Here, the baseline works well, giving both a good fit and effective decisions,
and the regularization term in the lookahead objective remains inactive.
In Figure \ref{fig:synth} (center), $\eta$ is larger.
Here, the baseline model pushes $x'$ points to a region where $y'$ values are low.
Meanwhile, the lookahead model, by incorporating into the objective
the decision model and estimating uncertainty surrounding $y'$,
is able to adjust the model to induce good decisions with some reduction in accuracy.
In Figure \ref{fig:synth} (right), $\eta$ is large.
Here, the $x'$ points are pushed far into areas of high uncertainty.
The success of  lookahead relies on the successful construction of intervals
at $\dist'$ through the successful estimation of $w$,
and may fail if $\dist$ and $\dist'$ differ considerably, as is the case here.

\begin{figure}[t]
	\centering
	\includegraphics[width=0.34\textwidth, clip]{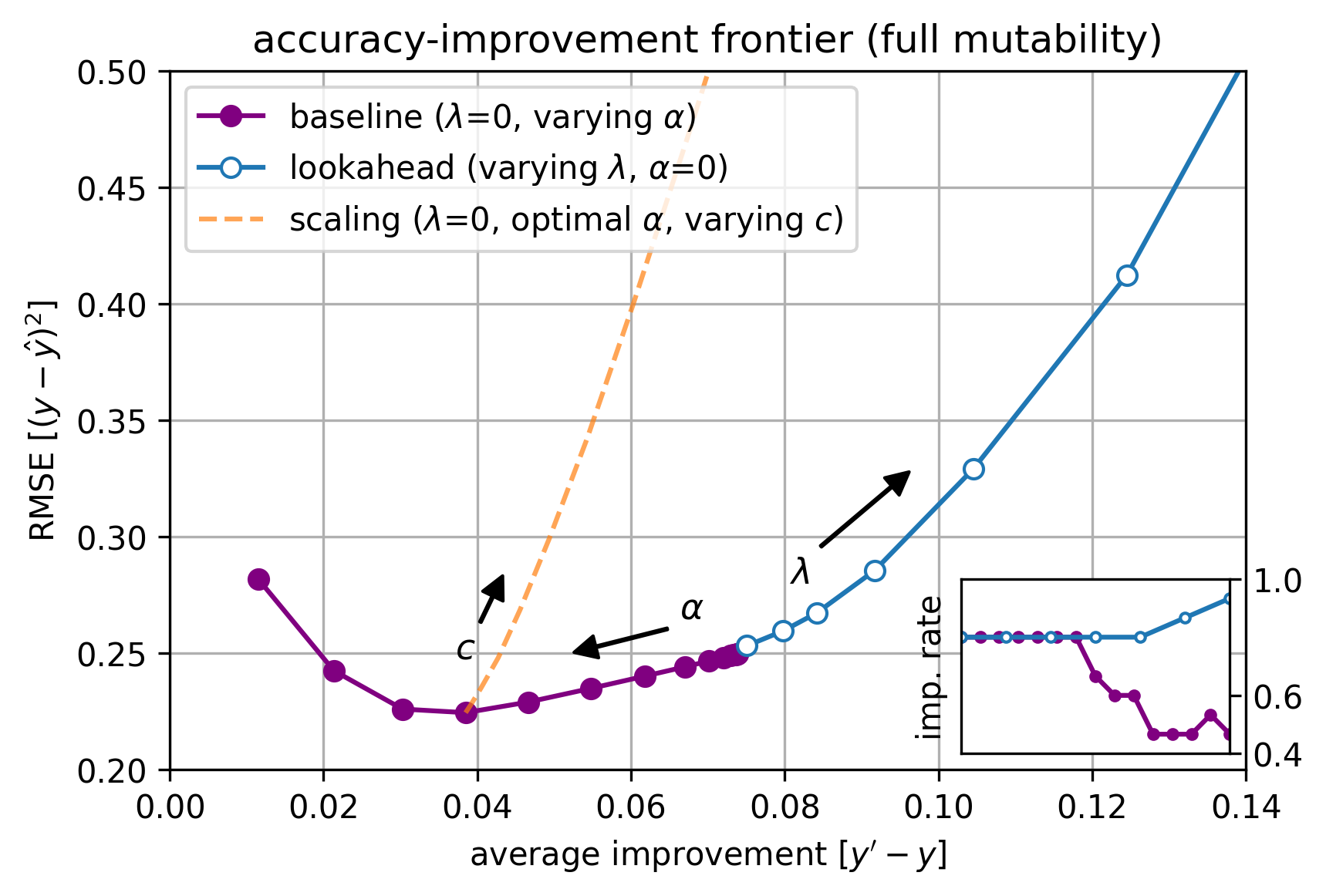}
	\includegraphics[width=0.34\textwidth, clip]{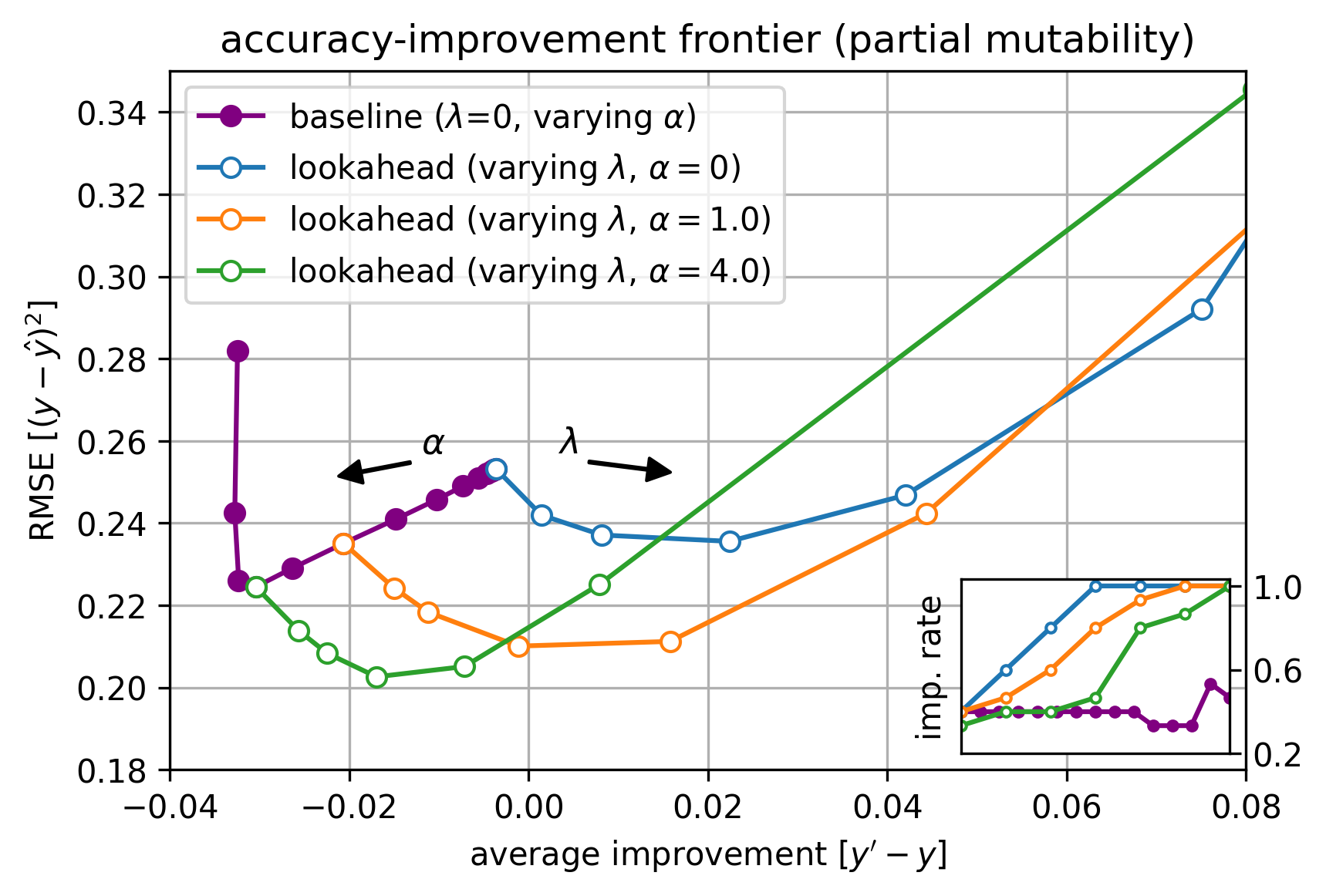}
	\includegraphics[width=0.29\textwidth, clip]{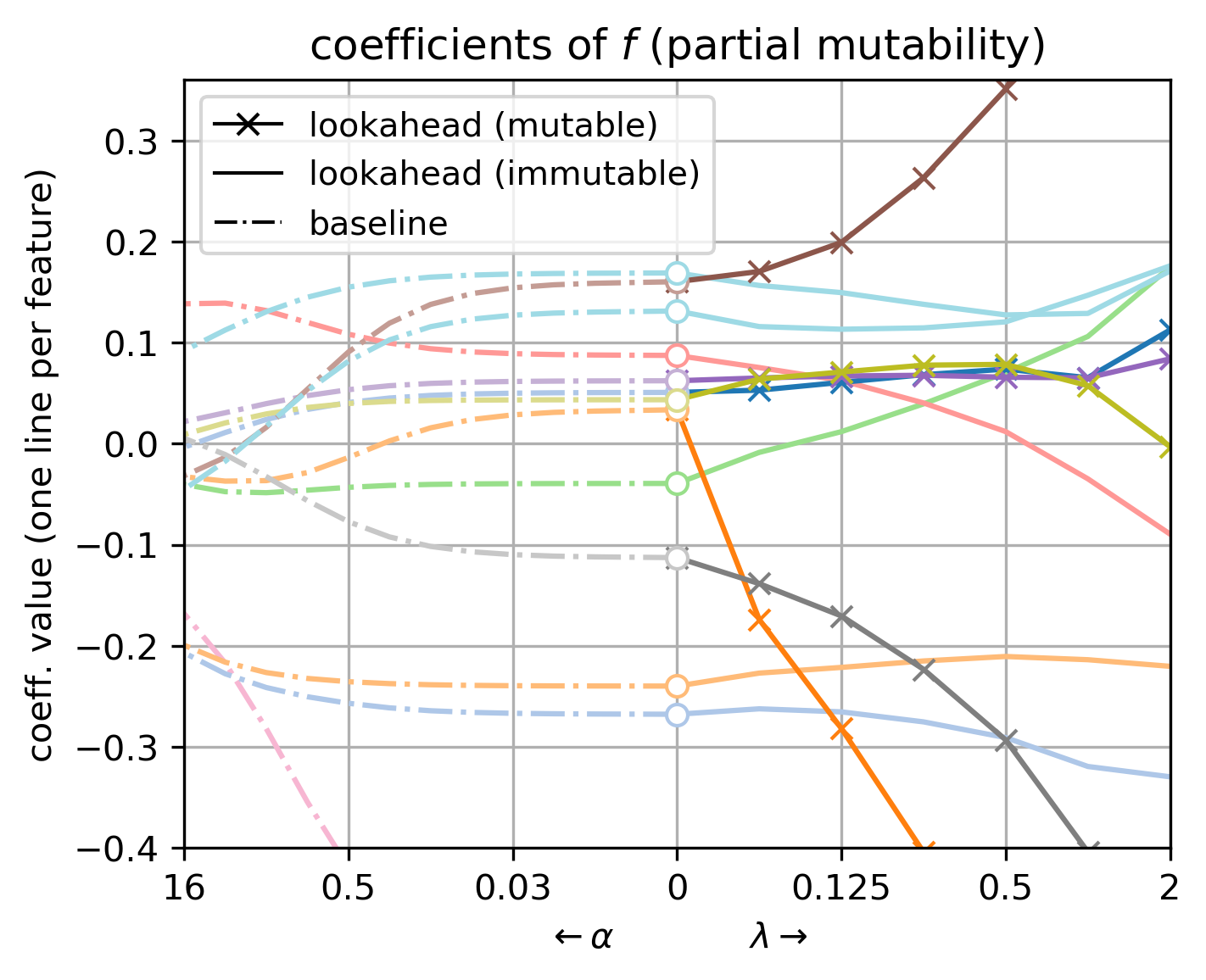}
	\caption{Results for the wine experiment. Tradeoff in accuracy and improvement under full mutability (left) and partial mutability (center),
	for which model coefficients are also shown (right).}
	\label{fig:wine}
\end{figure}

\subsection{Experiment 2: Wine quality}

The second experiment focuses on wine quality using the wine dataset from the UCI data repository \cite{dua2017uci}.
The wine in the data set has 13 features, most of which correlate linearly with quality $y$,
but two of which (alcohol and malic acid) have a non-linear U-shaped or inverse-U shaped relationship with $y$.
For the ground truth model, we set $f^*(x)=\sum_i \theta_i x_i + \sum_i \theta'_i x_i^2$ ($\text{RMSE}=0.2$, $y\in[0,3]$)
so that it captures these nonlinearities,
. To better demonstrate the capabilities of our framework, we 
sample points into the active set non-uniformly by thresholding on the non-linear features.
The active set includes $\sim$30\% of the data, and is further split 75-25
into a train set used for learning and tuning and a held-out test set used for final evaluation.

For the predictive model, our focus here is on linear models.
The baseline includes a linear $\fbase$ trained with $\ell_2$ regularization (i.e., Ridge Regression)
with regularization coefficient $\alpha\geq 0$.
Our lookahead model includes a linear $\fla$ 
trained with lookahead regularization (Eq. \eqref{eq:la_reg})
with regularization coefficient $\lambda\geq 0$. In some cases we will add to the objective an additional $\ell_2$ term,
so that for a fixed $\alpha$, setting $\lambda=0$ recovers the baseline model.
Lookahead was trained for 10 rounds and the baseline with a matching
number of overall epochs.
The uncertainty model $g$ uses residuals-based bootstrapping with 20 linear sub-models.
The propensity model $h$ is also linear.
We consider two settings:
one where all features (i.e., wine attributes) are mutable and using
decision step-size $\eta=0.5$,
and another where only a subset of the features are mutable and using step-size $\eta=2$.

\paragraph{Full mutability.}
Figure \ref{fig:wine} (left) presents the frontier of accuracy vs. improvement 
on the test set when all features are mutable.
The baseline and lookahead models coincide when $\alpha=\lambda=0$.
For the baseline, as $\alpha$ increases,
predictive performance (RMSE) displays a typical learning curve
with accuracy improving until reaching an optimum at some intermediate value of $\alpha$.
Improvement, however, monotonically decreases with $\alpha$,
and is highest with no regularization ($\alpha=0$).
This is because in this setting, gradients of $\fbase$ induce reasonably good decisions:
$\fbase$ is able to approximately recover the dominant linear coefficients of $f^*$,
and shrinkage due to higher $\ell_2$ penalization reduces the magnitude of the (typically positive, on average) change.
With lookahead, increasing $\lambda$ leads to better decisions,
but at the cost of higher (albeit sublinear) RMSE.
The initial improvement rate at $\lambda=0$ is high,
but  lookahead and $\ell_2$ penalties have opposing effects on the model.
Here, improvement is achieved by (and likely requires) increasing the 
size of the coefficients of linear model, $\fla$. We see that $\fla$ learns  to do this in an efficient way,
as compared to a {\naive} scaling of the predictively-optimal $\fbase$.

\paragraph{Partial mutability.}
Figure \ref{fig:wine} (center) presents the frontier of accuracy vs. improvement 
 when only a subset of the features are mutable (note that this effects the scale of possible improvement).
The baseline presents a similar behavior to the fully-mutable setting,
but with the optimal predictive model inducing a negative improvement.
Here we consider lookahead with various degrees of additional $\ell_2$ regularization.
When $\alpha=\lambda=0$, the models again coincide.
However, for larger $\lambda$, significant improvement can be gained with very little or no loss in RMSE,
while moderate $\lambda$ values improve both decisions and accuracy.
This holds for various values of $\alpha$, and setting $\alpha$ to the optimal value of $\fbase$
results in lookahead dominating the trade-off curve for all observed $\lambda$.
Improvement is reflected in magnitude and rate, which rises quickly from the baseline's $\sim 40\%$
to an optimal $100\%$, showing how lookahead learns models that lead to safe decisions. 

Figure \ref{fig:wine} (right) shows how the coefficients of
$\fbase$ and $\fla$ change as $\alpha$ and $\lambda$ increase, respectively (for lookahead $\alpha=0$).
As can be seen, lookahead works by making substantial changes to mutable coefficients,
sometimes reversing their sign, with milder changes to immutable coefficients. 
Lookahead achieves improvement by capitalizing on its freedom to
learn a useful direction of improvement within the mutable subspace,
while compensating for the possible loss in accuracy through mild changes in the immutable subspace.

\begin{figure}[t]
	\centering
	\includegraphics[width=0.34\textwidth, clip]{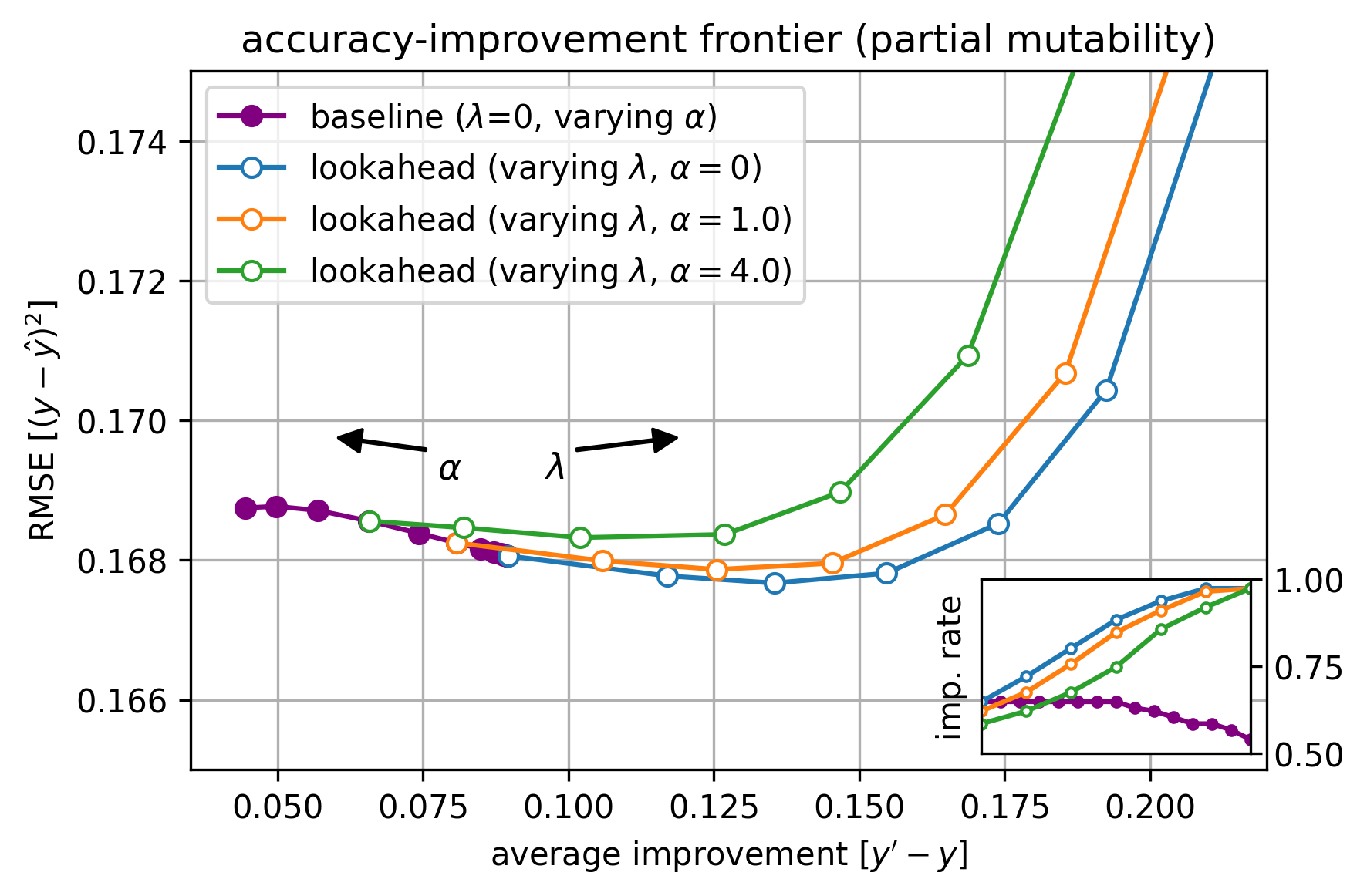}
	\includegraphics[width=0.315\textwidth, clip]{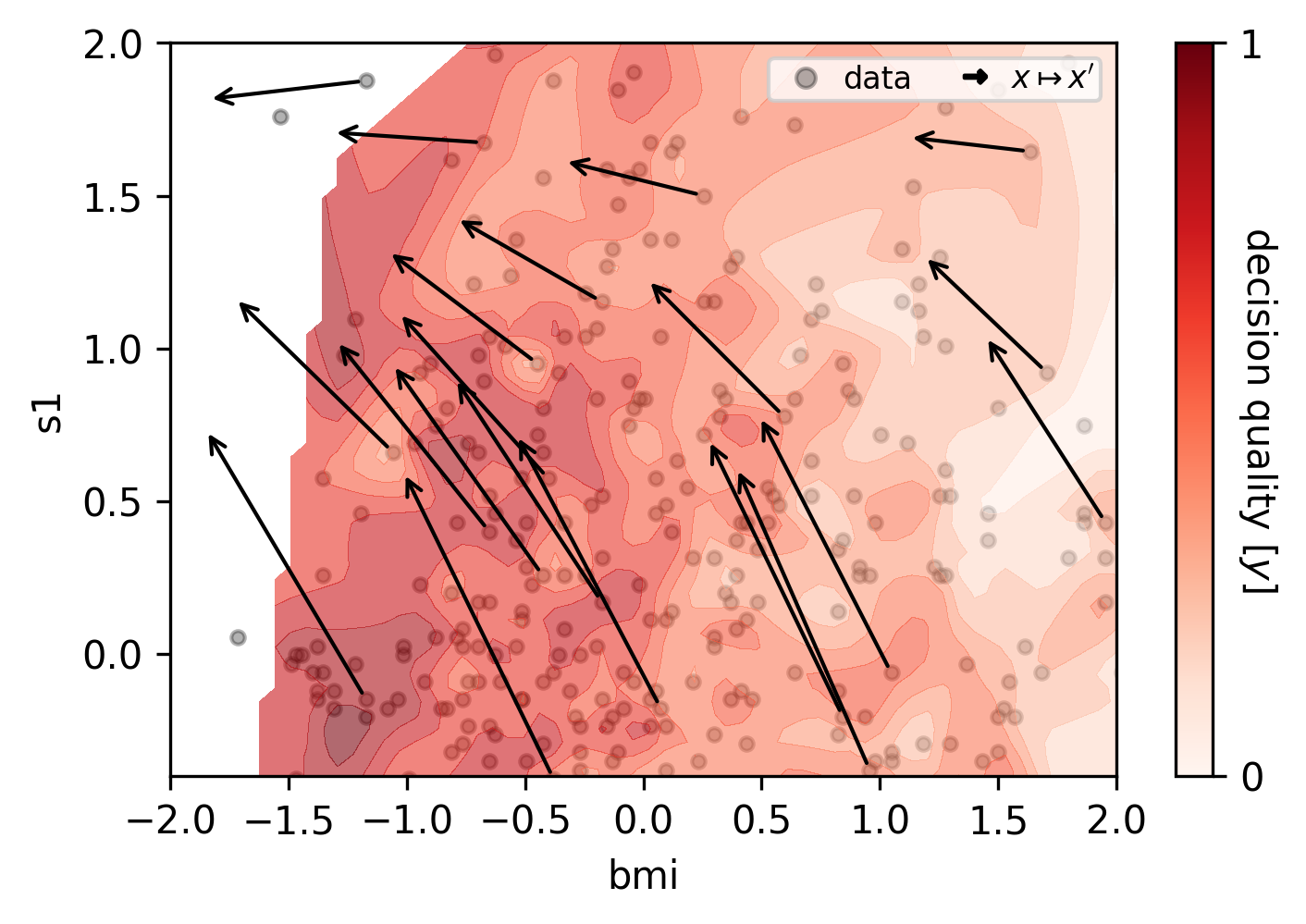}
	\includegraphics[width=0.315\textwidth, clip]{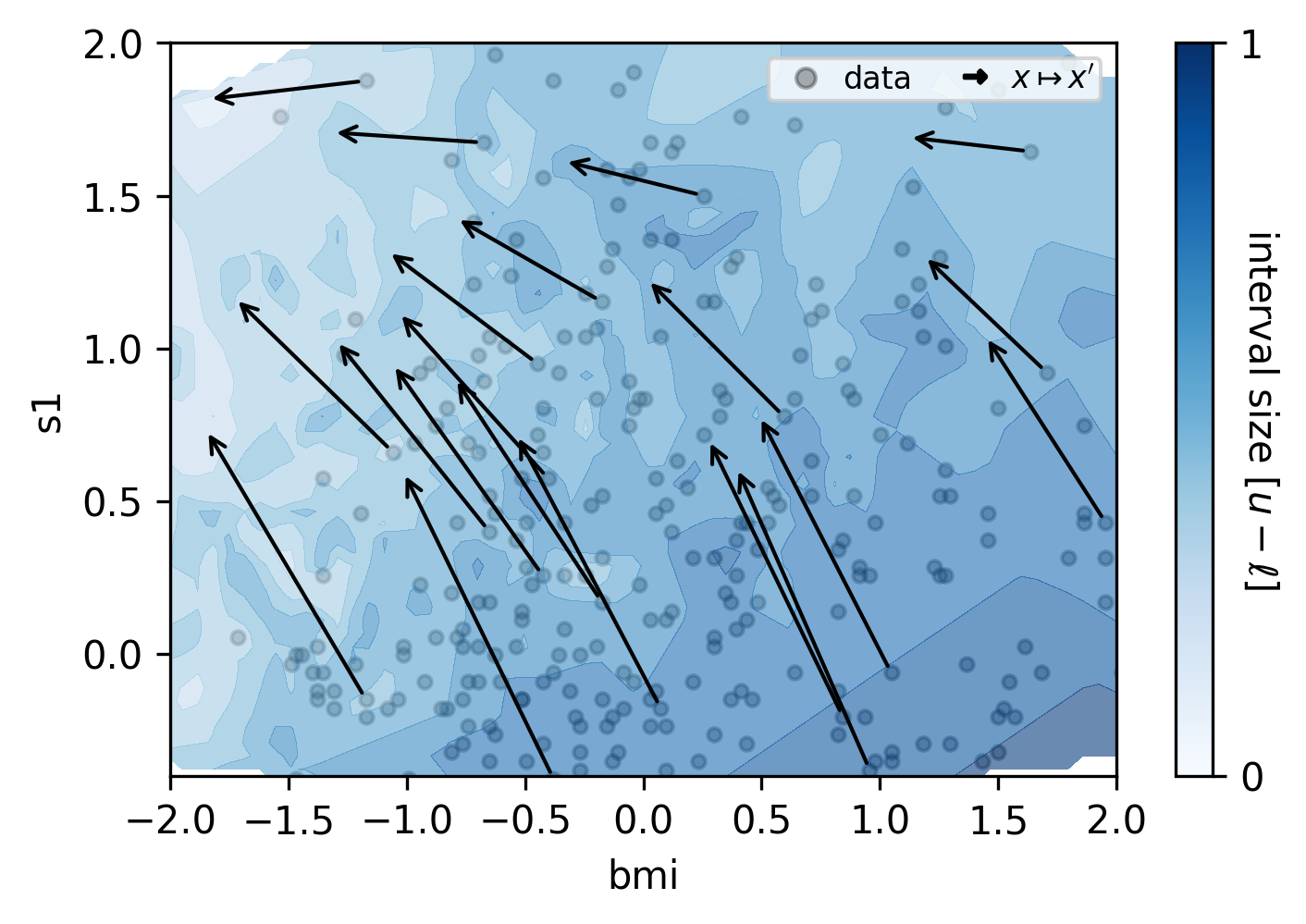}
	\caption{Results for the diabetes experiment. Tradeoff in accuracy and improvement under linear $f$ with partial mutability (left), visualization of shift $p \rightarrow p'$ with non-linear $f$ to regions of higher decision quality  (center), and regions of lower uncertainty (right).}
	\label{fig:diabetes}
\end{figure}

\subsection{Experiment 3: Diabetes}

The  final experiment focuses on the prediction of diabetes progression using the diabetes dataset\footnote{\texttt{https://www4.stat.ncsu.edu/\textasciitilde boos/var.select/diabetes.html}}  \cite{efron2004least}.
The dataset has 10 features describing various patient attributes.
We consider two features as mutable: BMI and T-cell count (marked as `s1').
While both display a similar (although reversed) linear relationship with $y$, feature s1 is much noisier.
The setup is as in wine but with two differences:
to capture nonlinearities we set $f^*$ to be a flexible generalized additive model (GAM) with splines of degree 10
($\text{RMSE}=0.15$),
and train and test sets are sampled uniformly from the data.
We normalize $y$ to $[0,1]$ and set $\eta=5$.

Figure \ref{fig:diabetes} (left) presents the accuracy-improvement frontier
for linear $f$ and bootstrapped linear $g$.
Results show a similar trend to the wine experiment,
with lookahead providing improved outcomes (both rate and magnitude) while preserving predictive accuracy.
Here,
lookahead improves results by learning to increase the coefficient of s1,
while adjusting other coefficients to maintain reasonable uncertainty.
The baseline fails to utilize s1 for improvement since from a predictive perspective
there is little value in placing weight on s1.

When $f$ is linear, decisions are uniform across the population in that
$\nabla_{f_\theta}(x)=\theta$ is independent of $x$.
To explore individualized actions, 
we also consider a setting where $f$ is a more flexible
quadratic model (i.e., linear in $x$ and $x^2$) in which gradients depend on $x$
and uncertainty is estimated using quantile regression.
Figure \ref{fig:diabetes} (center) shows the data as projected onto the subspace $(x_{\text{BMI}},x_{\text{s1}})$,
with color indicating outcome values $f^*(x)$,
interpolated within this subspace.
As can be seen, the mapping $x \mapsto x'$ due to $\fla$ generally improves outcomes.
The plot reveals that, had we had knowledge of $f^*(x)$,
uniformly decreasing BMI would also have improved outcomes,
and this is in fact the strategy envoked by the linear $\fbase$.
But decisions must be made based on the sample set, and so uncertainty must
be taken into account.
Figure \ref{fig:diabetes} (right) shows a similar plot but with color indicating uncertainty estimates as measured by the interval sizes given by $g$. 
The plot shows that decisions are directed towards regions of lower uncertainty
(i.e., approximately following the negative gradients of the uncertainty slope),
showing how lookahead successfully utilizes these uncertainties to adjust the predictive model $\fla$.


\section{Discussion}

Given the extensive use of machine learning across an ever-growing range of applications, we think it is appropriate to assume, as we have here, that 
predictive models will remain in widespread use,
and that at the same time, 
and despite well-understood concerns,
users will continue to act upon them.
In line with this, 
our goal with this work has been  to develop a machine learning framework that accounts for 
decision making by users
but remains fully within the discriminative framing of statistical machine learning.
The lookahead regularization framework that we  have proposed augments existing machine learning methodologies
with a component that promotes good human decisions.
We have demonstrated the utility of this approach across three different experiments, one
on synthetic data, one on predicting and deciding about wine, 
and one on predicting and deciding in regard to 
diabetes progression.
We hope that this work will inspire continued research in the machine
learning community that embraces predictive modeling while also being
cognizant of the ways in which our models are used.

\section*{Broader Impact}

In our work, the learning objective was designed to align with and support the possible use of a predictive model to drive decisions by users. 
It is our belief that a responsible and transparent deployment of
models with ``lookahead-like" regularization components should
avoid the kinds of mistakes that can be made when predictive methods
are conflated with causally valid methods.

At the same time, we have made a strong simplifying assumption, that of 
covariate shift, which requires that the relationship between covariates
and outcome variables is invariant as decisions are made and
the feature distribution changes. This strong assumption
is made to ensure validity for the lookahead regularization, since
we need to be able to perform inference about counterfactual 
observations. As discussed by~\citet{mueller2016learning} and~\citet{peters2016causal}, there exist real-world tasks that
reasonably satisfy this assumption, and yet at the same time, other
tasks--- notably those with unobserved confounders ---where this
assumption would be  violated. 
Moreover, this assumption is not testable on
the observational data.
This, along with the need to make an assumption about
the user decision model, means that an application of the
method proposed here should be done with care and will require
some domain knowledge to understand whether or not the assumptions
are plausible.

Furthermore, the validity of the interval estimates requires that any assumptions for the interval model used are satisfied and that weights $w$ provide a reasonable estimation of $\dist'/\dist$. In particular, fitting to $\dist'$ which has little to no overlap with $\dist$ (see Figure~\ref{fig:synth}) may result in underestimating the possibility of bad outcomes. 

 If used carefully and successfully, then the system provides
safety and protects against the misuse of a model. If used in
a domain for which the assumptions fail to hold then the framework
could make things worse, by trading accuracy for an incorrect view
of user decisions and the effect of these decisions on outcomes. 

We would also caution against any specific interpretation of the application of the model to the wine and diabetes data sets. We note that model misspecification of $f^*$ could result in arbitrarily bad outcomes, and estimating $f^*$ in any high-stakes setting requires substantial domain knowledge and should err on the side of caution. 
We use the data sets for purely illustrative purposes because we believe the
results are representative of the kinds of results that are available
when the method is correctly applied to a domain of interest.

\bibliographystyle{plainnat}
\bibliography{lookahead}

\clearpage

\appendix

\begin{appendices}

\section{Pseudocode}
Our algorithm alternates between optimizing the three components of the framework:
a predictive model, a propensity model, and an uncertainty model.
Here we give pseudocode for
the following per-component objectives:
\begin{enumerate}
\item A predictive model $\yhat=f(x)$, optimizing the squared loss:
\[\losspred(f;\smplst) = \sum_{i=1}^m (y_i-\yhat_i)^2\]
\item A propensity weight model $w=e^{h(x)}$, optimizing the log-loss:
\[\lossprop(h;\smplst,\smplst') = \sum_{i=1}^m
\log(1+e^{h(x_i)}) + \log(1+e^{-h(x'_i)}) \]
\item An uncertainty interval model $[\ell,u]=g_\conf(x)$, optimizing the $\conf$-quantile loss:
\[\lossuncert{\conf}(g;\smplst,w) = \sum_{i=1}^m w(x_i)
\max\{ (\conf-1)(y_i-\ell_i), \conf(y_i-\ell_i) \}\]
\end{enumerate}
but note that others can be plugged in. The pseudocode is given below.

\begin{algorithm}
\caption{Lookahead$(\smplst, T, \lambda, \eta, \conf)$}\label{algo:lookahead}
\begin{algorithmic}[1]
\State $f^{(0)} \gets \argmin_{f \in F} \losspred(f;\smplst)$
\For{$t=1,\dots,T$} 
\State $x'_i \gets \dec_\eta(x_i;f^{(t-1)})$ for all $i=1,\dots,m$ \Comment{e.g., $\dec_\eta(x;f)=x+\eta\mask(\nabla_f(x))$}
\State $\smplst' \gets \{x'_i\}_{i=1}^m$
\State $h^{(t)} \gets \argmin_{h \in H} \lossprop(h;\smplst,\smplst')$
\State $w \gets e^{h^{(t)}}$ 
\State $g^{(t)} \gets \argmin_{g \in G} \lossuncert{\conf}(g;\smplst,w)$
\State $f^{(t)} \gets \argmin_{f \in F} \losspred(f;\smplst) +
\lambda \reg(g^{(t)};\smplst)$
\EndFor
\State \textbf{return} $f^{(T)}$
\end{algorithmic}
\end{algorithm}

\section{Uncertainty models}
Here we describe the two uncertainty methods used in our paper
and how they apply to our setting.

\subsection{Bootstrapping}
Bootstrapping produces uncertainty intervals by combining
the outputs of a collection of $k$ models $\{g^{(i)}\}_{i=1}^k$,
each trained independently for \emph{prediction} on a random subset of the data.
There are many approaches to bootstrapping, and here we describe two:
\begin{itemize}
\item \textbf{Vanilla bootstrapping}:
Each $g^{(i)}$ is trained using a predictive objective (e.g., squared loss)
on a sample set $\smplst^{(i)}=\{(x^{(i)}_j,y^{(i)}_j)\}_{j=1}^m$ where
$(x^{(i)}_j,y^{(i)}_j)$ are sampled with replacement from $\smplst$.
The sub-models are then combined using:
\[
g(x) = [\mu(x)-z\sigma(x), \mu(x)+z\sigma(x)]
\]
where:
\[
\mu(x) = \frac{1}{k}\sum_{i=1}^k g^{(i)}(x), \quad\qquad
\sigma(x) = \frac{1}{k}\sum_{i=1}^k (\mu(x) - g^{(i)}(x))^2
\]
and $z$ is the z-score corresponding to the confidence parameter $\conf$
under a normal distribution.

\item \textbf{Bootstrapping residuals}:
First, a predictive model $\bar{g}$ is fit to the data,
and residuals $r=y-\bar{g}(x)$ are computed.
Then, each $g^{(i)}$ is trained on the original sample data
but with ground truth-labels $y_i$ replaced with random pseudo-labels:
\[
\smplst^{(i)}=\{(x_j,\ybar^{(i)}_j)\}_{j=1}^m \qquad \quad
\ybar^{(i)}_j = y_i+r_j
\]
where $r_j$ are sampled with replacement from $\{r_j\}_{j=1}^m$.
\end{itemize}

In our framework, because $g$ must apply to $\dist'$,
each $g^{(i)}$ is trained with propensity weights $w$.
To account for cases where $\dist$ and $\dist'$ differ,
the $g^{(i)}$ are trained not on sample sets of size $m$,
but rather, of size $\tilde{m}(w)$,
where $\tilde{m}(w)$ is the \emph{effective sample size} \cite{kong1994sequential} given by:
\[
\tilde{m}(w) =
\frac{\mathrm{mean}(\{w_i\}_{i=1}^m)}{\mathrm{var}(\{w_i\}_{i=1}^m)},
\qquad w_i=w(x_i) \,\,\, \forall \, i=1,\dots,m
\]

\subsection{Quantile regression}
Quatile regression is a learning framework for training models
to predict the $\conf$-quantile of the conditional label distribution $\dist(y|x)$.
Just as training with the squared loss is aimed at predicting the mean
of $\dist(y|x)$, training with the absolute loss 
$|y-\yhat|$ is aimed at the median.
Quantile regression generalizes the absolute loss by considering a 'tilted'
variant with slopes $\conf-1$ and $\conf$:
\[
Q_\conf(y,\yhat) = \max\{(1-\conf)(y-\yhat), \conf (y-\yhat)\}
\]

\section{Experimental details}

\subsection{Experiment 1: Quadratic curves}
Here we set $f^*(x)=-0.8x^2+0.5x+0.1$.
$F$ and $G$ include quadratic functions,
and $H$ to include linear functions. For uncertainty estimation we used
vanilla bootstrap, and for propensity scores we used logistic regression.
For lookahead, we set $\lambda=4$, $\conf=0.95$,
use $k=10$ bootstrapped models, and train for $T=5$ rounds.
The data includes $m=25$ samples $x$ drawn from $N(-0.8,0.5)$,
and $y=f^*(x)+\epsilon$ where $\epsilon \sim N(0, 0.25)$.
We use a $75:25$ train-test split.
The three conditions vary only in $\eta$
with values $\eta=0.75, 1.25$, and $3.5$.

Quantitative results are given in the table below:
\begin{table}[h!]
  \centering
    \begin{tabular}{clrrr}
      &   & \multicolumn{1}{l}{RMSE} & \multicolumn{1}{l}{Imp. rate} & \multicolumn{1}{l}{Imp. mag.} \bigstrut[b]\\
    \hline
    \multirow{2}[2]{*}{$\eta=0.75$} & baseline & 0.349 & 0.857 & 1.109 \bigstrut[t]\\
      & lookahead & 0.351 & 0.857 & 1.108 \bigstrut[b]\\
    \hline
    \multirow{2}[2]{*}{$\eta=1.25$} & baseline & 0.342 & 0.143 & -0.261 \bigstrut[t]\\
      & lookahead & 0.424 & 0.714 & 1.065 \bigstrut[b]\\
    \hline
    \multirow{2}[1]{*}{$\eta=3.5$} & baseline & 0.342 & 0 & -35.13 \bigstrut[t]\\
      & lookahead & 0.675 & 0.571 & 0.604 \\
    \end{tabular}%
  \label{tbl:synth}%
\end{table}%

\subsection{Experiment 2: Wine quality}
The wine dataset includes $m=178$ examples and $d=13$ features.
We learn a quadratic $f^*(x)=\sum_i \theta_i x_i + \sum_i \theta'_i x_i^2$.
$F$, $G$, and $H$ include linear functions.
For uncertainty estimation we used residuals bootstrap, and for propensity scores we used logistic regression.
For lookahead, we set $\conf=0.95$, use $k=20$ bootstrapped models, and train for $T=10$ rounds.
For $f$, we use SGD with a learning rate of 0.1 and 1000 epochs for initialization
and 100 additional epochs per round.
For $g$, each sub-model was trained with SGD using a learning rate of 0.1 and for 500 epochs.
We set $\eta=0.5$ and $\eta=2$ for the fully and partially mutable settings, respectively.

\subsection{Experiment 3: Diabetes}
The diabetes dataset includes $m=442$ examples and $d=10$ features.
We set $f^*(x)$ to be a generalized additive model (GAM) with splines of degree 10
trained on the entire dataset and tuned using cross-validation.
In the first setting, $F$, $G$, and $H$ include linear functions.
In the second setting, $F$, $G$ are quadratic functions (i.e., linear in $x_i$ and in $x^2_i$)
and $H$ remains linear.
For uncertainty estimation we used quantile regression, and for propensity scores we used logistic regression.
For lookahead, we set $\conf=0.8$ and train for $T=10$ rounds.
For $f$, we use SGD with a learning rate of 0.05 and 1000 epochs for initialization
and 100 additional epochs per round.
For $g$, we use SGD with a learning rate of 0.05 and for 500 epochs.
For both linear and non-linear settings we set $\eta=5$, and normalize $y$ to be in $[0,1]$.

%
%
%
%
%

\end{appendices}

\end{document}